\pdfoutput=1
\documentclass[11pt]{article}
\usepackage[table,dvipsnames]{xcolor}
\usepackage{acl}
\usepackage{times}
\usepackage{latexsym}
\usepackage[T1]{fontenc}
\usepackage[utf8]{inputenc}
\usepackage{microtype}
\usepackage{inconsolata}
\usepackage{amsmath}
\usepackage{amssymb}
\usepackage{graphicx}
\usepackage{enumerate}
\usepackage{balance}
\usepackage{blindtext}
\usepackage{booktabs}
\usepackage{colortbl}
\usepackage{float}
\usepackage{makecell}
\usepackage{multicol}
\usepackage{multirow}
\usepackage{tabularx}
\usepackage{marvosym}
\usepackage{latexsym}
\usepackage{listings}
\usepackage{tcolorbox}
\usepackage{stfloats}
\usepackage{subfigure}
\usepackage{subcaption}
\usepackage{ulem}
\lstdefinestyle{PythonCode}{
    language=Python,
    basicstyle=\ttfamily,
    breaklines=true,
    keywordstyle=\bfseries\color{NavyBlue},
    morekeywords={},
    emph={self},
    emphstyle=\bfseries\color{Rhodamine},
    commentstyle=\itshape\color{black!50!white},
    stringstyle=\bfseries\color{PineGreen!90!black},
    columns=flexible,
}

\lstdefinestyle{BashCode}{
    language=Bash,
    basicstyle=\ttfamily\color{white},
    backgroundcolor=\color{black},
    breaklines=true,
    keywordstyle=\bfseries\color{MidnightBlue},
    morekeywords={},
    emph={},
    emphstyle=\bfseries\color{Purple},
    commentstyle=\itshape\color{black!50!white},
    stringstyle=\bfseries\color{OliveGreen!90!black},
    columns=flexible,
}

\newcommand{\ie}{\textit{i.e., }}
\newcommand{\eg}{\textit{e.g., }}

\newcommand\footnoteONLYtext[1]{
    \let \mybackup \thefootnote
    \let \thefootnote \relax
    \footnotetext{#1}
    \let \thefootnote \mybackup
    \let \mybackup \imareallyundefinedcommand}

\title{Iterative Experience Refinement of Software-Developing Agents}

\author{
  \textbf{Chen Qian}$^{\dagger\spadesuit}$ \quad
  \textbf{Jiahao Li}$^{\dagger\bigstar}$ \quad 
  \textbf{Yufan Dang}$^\spadesuit$ \quad 
  \textbf{Wei Liu}$^\spadesuit$ \quad
  \textbf{YiFei Wang}$^\spadesuit$ \quad
  \textbf{Zihao Xie}$^\spadesuit$ \quad \\
  \textbf{Weize Chen}$^\spadesuit$ \quad 
  \textbf{Cheng Yang}$^\clubsuit$\textsuperscript{\Letter} \quad 
  \textbf{Yingli Zhang}$^\blacklozenge$ \quad
  \textbf{Zhiyuan Liu}$^{\spadesuit}$\textsuperscript{\Letter} \quad  
  \textbf{Maosong Sun}$^{\spadesuit}$\textsuperscript{\Letter} \\
  $^\spadesuit$Tsinghua University \quad
  $^{\bigstar}$Dalian University of Technology \\
  $^\clubsuit$Beijing University of Posts and Telecommunications \quad
  $^\blacklozenge$Siemens \\
  \texttt{qianc62@gmail.com} \quad 
  \texttt{lijihao2021@mail.dlut.edu.cn} \quad \\
  \texttt{yangcheng@bupt.edu.cn} \quad 
  \texttt{liuzy@tsinghua.edu.cn} \quad 
  \texttt{sms@tsinghua.edu.cn}
}

\begin{document}

\maketitle

\footnoteONLYtext{$^\dagger$Equal Contribution.}
\footnoteONLYtext{$^{\text{\Letter}}$Corresponding Author.}

\begin{abstract}
Autonomous agents powered by large language models (LLMs) show significant potential for achieving high autonomy in various scenarios such as software development. Recent research has shown that LLM agents can leverage past experiences to reduce errors and enhance efficiency. However, the static experience paradigm, reliant on a fixed collection of past experiences acquired heuristically, lacks iterative refinement and thus hampers agents' adaptability.
In this paper, we introduce the \textit{Iterative Experience Refinement} framework, enabling LLM agents to refine experiences iteratively during task execution. We propose two fundamental patterns: the successive pattern, refining based on nearest experiences within a task batch, and the cumulative pattern, acquiring experiences across all previous task batches. Augmented with our heuristic experience elimination, the method prioritizes high-quality and frequently-used experiences, effectively managing the experience space and enhancing efficiency.
Extensive experiments show that while the successive pattern may yield superior results, the cumulative pattern provides more stable performance. Moreover, experience elimination facilitates achieving better performance using just 11.54\% of a high-quality subset.
\end{abstract}

\vspace{-12pt}
\section{Introduction}
Recently, large language models (LLMs) have attained remarkable success, showcasing substantial potential in approximating human-like intelligence~\cite{vaswani2017attention,NEURIPS2020_1457c0d6,bubeck2023sparks,Wang2023ASO}. Fueled by the vast progress of LLM, autonomous agents based on LLM have emerged, endowed with memory~\cite{park2023generative}, planning~\cite{wei2022chain}, and tool using~\cite{schick2023toolformer}. These enhancements elevate the capabilities of LLM-based autonomous agents, enabling them to adapt to more complicated scenarios~\cite{park2023generative,wang2023humanoid,hua2023war,wang2023voyager,wang2023avalons} and tackle a broader range of tasks~\cite{GPTEngineer,gong2023mindagent,qian2023communicative}. 
Furthermore, the advancement of LLM-based autonomous agents has brought about another significant breakthrough through the integration of multi-agent cooperation~\cite{park2023generative,li2023camel,qian2023communicative}. Through involvement in multi-turn communication, agents actively engage in responsive or instructive conversations, collaboratively improving the autonomous attainment of a cohesive solution for task completion. This paradigm fosters greater autonomy of the agent, consequently decreasing reliance on human engagement~\cite{li2023camel,qian2023communicative,wu2023autogen}. 
In order to study the cooperative dynamics of autonomous agents more pertinently, we choose software development as an representative scenario. This choice is motivated by its \textit{complexity}, which requires a combination of programming and natural language skills~\cite{mills1976software}, the \textit{processuality} that always demands an deep coding understanding and continuous adjustment~\cite{barki1993toward}, and the \textit{objectivity} of code that can provide quantifiable feedback~\cite{compton2002reconfigurable}.
An example is ChatDev~\cite{qian2023communicative}, where LLM-based autonomous agents play various roles (\eg an instructive reviewer and a responsive programmer) in a waterfall-like workflow, cooperatively participating in the software development process through their multi-turn communication.

Along this line, recent research has focused on enabling agents to efficiently learn from past experiences, aiming to prevent recurring errors and enhance efficiency in subsequent task execution~\cite{qian2023experiential}. 
These agents work collaboratively to acquire and leverage experiences acquired from past task executions, substantially enhancing agents' autonomy and their proficiency in collectively addressing unseen tasks.
However, the acquisition of experiences was a one-time process using heuristic rules~\cite{zhao2023expel,qian2023experiential}. This static experience paradigm restricts the agent's ability to adapt to intricate tasks such as software development, as it lacks the iterative refinement of experiences necessary for adaptive improvement.

To this end, we propose a novel \textit{Iterative Experience Refinement} (IER) framework, wherein agents iteratively refine their past experiences during task execution. This iterative process involves a cycle of continual acquisition and utilization of experiences. 
Technically, we establish two foundational patterns for experience refinement across various tasks: the \textit{successive pattern} and the \textit{cumulative pattern}. 
In the successive pattern, experiences are derived from the latest task batch, while the cumulative pattern integrates all historical experiences from previous task batches.
Moreover, the process of accumulating experiences may inadvertently lead to an undesired expansion of the experience space, inevitably encompassing numerous low-quality or rarely-used ones.
Correspondingly, we propose a heuristic \textit{experience elimination} mechanism, which prioritizes frequently employed experiences in task execution while discarding identified low-quality ones, streamlining the evolution of experiences toward greater efficiency.
We conducted experiments from the perspectives of both software quality and experience refinement, demonstrating the superior effectiveness of iterative refinement in enhancing agents' experiences for task execution.

In summary, the main contributions that we have made are outlined as follows:
\begin{enumerate}[$\bullet$]
\setlength{\parsep}{0pt}
\setlength{\topsep}{0pt}
\setlength{\itemsep}{0pt}
\item To the best of our knowledge, we are the first to introduce a novel experience refinement framework. This new paradigm, grounded in the dynamic iteration of past experiences, empowers agents to adaptively solve new tasks through continual acquisition, utilization and elimination.
\item We propose a heuristic mechanism to experience elimination that prioritizes high-quality and frequently-utilized experiences, thereby mitigating inefficiency issues arising from the potential expansion of the experience space.
\item Through extensive experiments, we found that while the successive pattern may yield higher results, the cumulative pattern offers more stable performance. Besides, experience elimination facilitates achieving better performance with just 11.54\% of a high-quality subset of experiences.
\end{enumerate}

\vspace{-10pt}
\section{Related Work}
Recent progress in LLMs makes it play a vital role in natural language processing ~\cite{NEURIPS2020_1457c0d6,bubeck2023sparks,vaswani2017attention,radford2019language,touvron2023llama,wei2022emergent,Shanahan2023,chen2021evaluating,brants2007large,chen2021evaluating,ouyang2022training,yang2023large,qin2023large,kaplan2020scaling,Shumailov2023TheCO} and further foster the development of autonomous agents in independently solving tasks ~\cite{zhou2023webarena,wang2023voyager,park2023generative,wang2023humanoid,AutoGPT,GPTEngineer,wang2023promptagent}. And enhanced with tool using ~\cite{schick2023toolformer,cai2023large,qin2023toolllm,ruan2023tptu,GPT4Tools}, memory~\cite{park2023generative,sumers2023cognitive} and planning~\cite{chen2023agentverse,liu2023bolaa}, autonomous agents utilize robust capability of LLMs in complementing complex tasks ~\cite{zhou2023webarena,ma2023laser,zhang2023generative,wang2023large,ding2023designgpt,weng2023prompt,GPTEngineer, park2023generative,zhou2023agents,zhu2023ghost}.

When involved in autonomous communication among multiple agents, these multi-agent systems exhibit enhanced capabilities for addressing complex tasks~\cite{li2023camel,qian2023communicative,wu2023autogen,hong2023metagpt,li2023metaagents,park2023generative,zhou2023agents,chen2023agentverse,chan2023chateval,chen2023gamegpt,Cohen2023LMVL,hua2023war}.
More recently, agents endowed with instructive and responsive experiences further exhibit notable promotion in their cooperative task execution~\cite{zhong2023memorybank,lewis2021retrievalaugmented}. 

For example, 
ExpeL~\cite{zhao2023expel} innovatively accumulates experiences from successful historical trajectories and leverages these experiences during inference. 
Experiential Co-Learning (ECL)~\cite{qian2023experiential} focuses on collecting shortcut-oriented experiences from past trajectories, enabling agents to more effectively handle unseen tasks.

\vspace{-6pt}
\section{Methodology}
The previous design of agents' experiences is through heuristic rules in a one-time process~\cite{qian2023experiential}; these static experiences lack the capability to be dynamically refined during future task executions. 
To tackle the challenge, we introduce a \textit{iterative experience refinement} (IER) framework, in which agents are equipped with experiences that undergo dynamic refinement during agents' task-solving processes. 
Given the overly fine-grained propagation of experiences between individual tasks, our approach involves partitioning all tasks into non-overlapping task batches \(\langle \mathcal{T}_1,\mathcal{T}_2,\cdots,\mathcal{T}_n \rangle\). In each batch, agents iteratively generate a new experience pool, which is then propagated to the subsequent batches.
In the following, similar to inheritance, experiences are generated and propagated from preceding task batches (\ie predecessors) to subsequent task batches (\ie descendants).
Technically, we have developed two primary patterns for the propagation of experience between various tasks: the successive pattern and the cumulative pattern. 
In the successive pattern, a descendant' experience pool inherits from the latest predecessor, prioritizing the latest knowledge, while in the cumulative pattern, a descendant inherits the historical experiences from all previous predecessors.
Furthermore, considering the practical limitation of the experience pool size, we propose empirically powerful mechanism for eliminating low-quality experiences during the propagation process based on the information density and utilization frequency of the agents' experiences.
\looseness=-1

\vspace{-6pt}
\subsection{Experience Acquisition and Utilization}

\begin{figure}[t]
    \centering
    \includegraphics[width=0.45\textwidth]{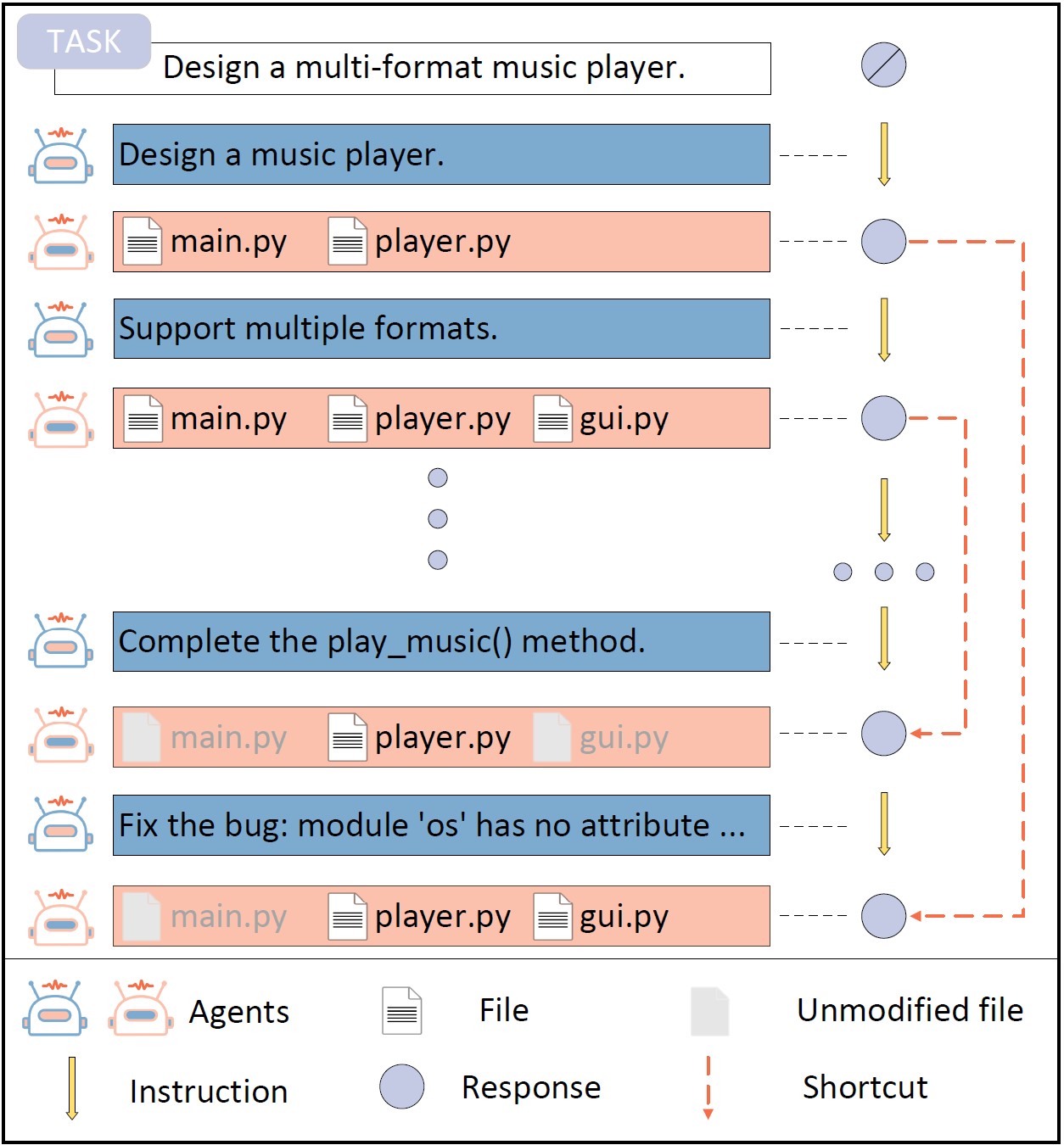}
    \caption{The task execution chain constructed for shortcut-oriented experience acquiring. The execution chain creates procedural trajectories for various training tasks, where we acquire "shortcuts" linking non-adjacent nodes as agents' experiences.}
    \label{fig:chain}
    \vspace{-10pt}
\end{figure}

This module aims to execute the tasks within every batch to acquire experiences for continuous iteration of the experience pool.
Illustrated in Figure~\ref{fig:chain}, following \citet{qian2023experiential} where an instructive agent and a responsive agent are involved, throughout their cooperative task execution, we record a series of instructions (\(\mathcal{I}\!=\!\{i_1, i_2, \cdots, i_n\}\)) alongside a corresponding series of responded solutions (\(\mathcal{S}\!=\!\{s_1, s_2, \cdots, s_n\}\)), with each solution representing a complete software code. The communication process is described as a directed chain  \( G = (N, {E}) \):
\begin{equation}
\begin{aligned}
N\!&=\!\{ s_j | s_j \in \mathcal{S} \}\!\cup\!\{s_0\} \\
{E}\!&=\!\{ (s_j, i_{j+1}, s_{j+1}) | s_j, s_{j+1}\!\in\!\mathcal{S}, i_{j+1}\!\in\!\mathcal{I} \}
\end{aligned}
\end{equation}
where \( N \) denotes the nodes that correspond to the solutions (with \( s_0 \) represents the initial, usually empty solution), and \( {E} \) represents the edges 
that correspond to the instructions. And each edge \( (s_j, i_{j+1}, s_{j+1}) \) is guided by the instruction \( i_{j+1} \), illustrating the transition from one solution \( s_j \) to the modified one \( s_{j+1} \).

\vspace{-6pt}
\paragraph{Acquisition}
Recognizing that not all progressions in the chain (\ie a single round of software optimization) necessarily lead to better solutions, we opt to acquire more efficient experiences from non-existing edges in the chain. As depicted in Figure~\ref{fig:chain}, we traverse non-adjacent node pairs along the execution chain and acquire all "shortcuts" linking non-adjacent nodes:
\begin{equation}
\begin{aligned}
\mathcal{E}=\{ (s_i, \overset{\dashrightarrow}{s_is_j}, s_j) | s_i, s_j\!\in\!N\wedge\\\ \ (s_i, \cdot, s_j)\notin\mathcal{E} \wedge [\![ s_i\rightarrow s_j ]\!] \}
\end{aligned}
\end{equation}
where $[\![ s_i\!\rightarrow\!s_j ]\!]$ indicates that $s_j$ is reachable from $s_i$, \(\overset{\dashrightarrow}{s_is_j}\) denotes a pseudo instruction via a standard self-instruction mechanism~\cite{wang-etal-2023-self-instruct}.
This mechanism extracts shortcuts instead of the existing edges, which can effectively motivate agents use the shortcuts as their experiences to engage in shortcut thinking and has been validated by previous work~\cite{qian2023experiential}. 

\vspace{-6pt}
\paragraph{Utilization}
After experience acquisition, a shortcut $(s_i, \overset{\dashrightarrow}{s_is_j}, s_j)$ can be divided into $(s_i, \overset{\dashrightarrow}{s_is_j})$, indicating solution-to-instruction knowledge possessed by the instructive agent, and $(\overset{\dashrightarrow}{s_is_j}, s_j)$, signifying instruction-to-solution knowledge held by the responsive agent. These distinct key-value forms of knowledge are amalgamated into the agents' experience pools for utilization in their collective reasoning.
When executing a unseen task, for a current solution \( s_j \), the instructive agent initially employs retrieval mechanism to access empirical instructions closely matching the latent meaning of the query from the solution-to-instruction experience pool. These retrieved instructions serve as few-shot examples to assist in generating an experience-augmented instruction \( i_{j+1}^* \). Similarly, the responsive agent retrieves the responses with the highest matching degree to the instruction from the instruction-to-solution experience pool, serving as few-shot examples to assist in responding with an experience-augmented solution \( s_{j+1}^* \).
Thus, the whole reasoning process is represented as a sequence of pairs $\{(i_1^*, s_1^*), (i_2^*, s_2^*), \cdots\}$, where each includes an experience-enhanced instruction and the corresponding solution.

\vspace{-6pt}
\subsection{Experience Propagation}
Nonetheless, these static experiences can limit agents' adaptability to new tasks and hinder continuous learning. Addressing the rigidity of experiences is essential for overcoming these limitations.
As agents accumulate experiences from predecessors for use in the current batch, the current batch can also naturally generate new experiences that are propagated for descendants.
Based on this, we propose two types of fundamental experience refinement patterns, namely the successive pattern and the cumulative pattern.

\begin{figure}[t]
    \centering
    \includegraphics[width=0.40\textwidth]{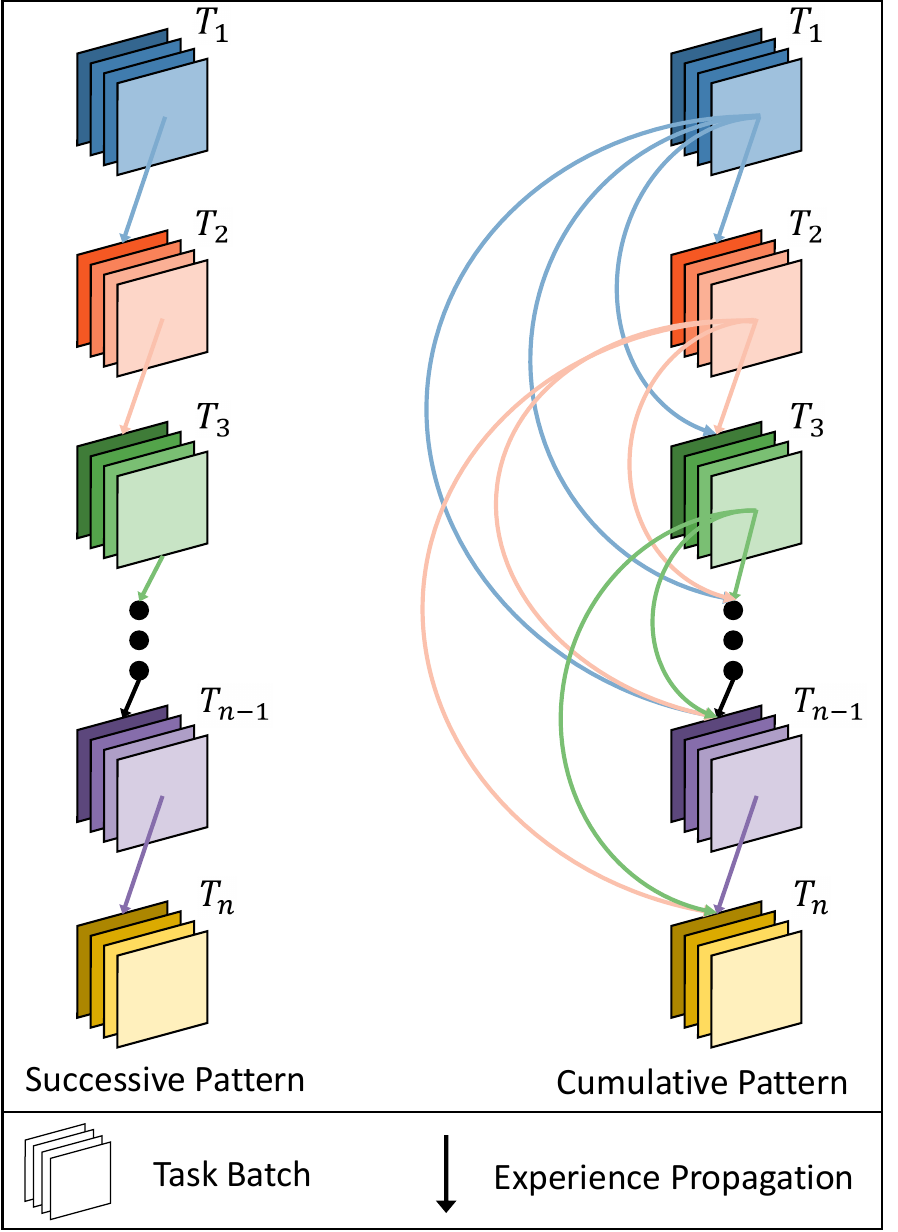}
    \caption{The successive pattern (left) allows each task batch to utilize the experience pool collected from the preceding batch. The cumulative pattern (right) enables each batch of tasks to leverage the experience pool acquired from all previous batches.}
    \label{fig:pattern}
    \vspace{-10pt}
\end{figure}

\vspace{-6pt}
\paragraph{Successive Pattern} Leveraging recently acquired experiences aligns naturally with our objectives. Inspired by this insight, we introduce the successive pattern, depicted on the left side of Figure~\ref{fig:pattern}. When executing a task batch \( \mathcal{T}_i \), agents can gather experiences \(\mathcal{E}_{i-1}\) acquired from the nearest predecessor, \ie \( \mathcal{T}_{i-1} \), which constitutes their experiences in the next descendant.
Using $\mathcal{E}$ to represent the acquired experience pool, and $\mu(\mathcal{E},\mathcal{T})$ to denote experience utilization on a task batch $\mathcal{T}$, this process can be expressed as follows:
\begin{equation}
\begin{aligned}
&\mathcal{E}_{1} = \emptyset, \quad \mathcal{E}_{i} = \mu(\mathcal{E}_{i-1}, \mathcal{T}_i) \\
\end{aligned}
\end{equation}

\vspace{-6pt}
\paragraph{Cumulative Pattern} Alternatively, we explore whether continuous experience accumulation can elevate task-solving abilities. In the cumulative pattern illustrated on the right side of Figure~\ref{fig:pattern}, agents can employ experiences from all previous experience pools \(\{ \mathcal{E}_1, \mathcal{E}_2, ..., \mathcal{E}_{i-1} \}\) in the execution of the task batch \( \mathcal{T}_i \):
\begin{equation}
\begin{aligned}
&\mathcal{E}_{1} = \emptyset, \quad \mathcal{E}_{i} = \mu \big( \bigcup_{j=1}^{i-1}\mathcal{E}_j, \mathcal{T}_i \big) \\
\end{aligned}
\end{equation}
These two types of experience propagation can be likened to the passing down of knowledge through generations. The former is akin to descendant inheriting knowledge from their parents, while the latter is akin to descendant inheriting knowledge from their parents and all previous predecessors.

\vspace{-6pt}
\subsection{Experience Elimination}
Recognizing that the process of accumulating experiences may inadvertently lead to an undesired expansion of the experience space, inevitably encompassing numerous low-quality or rarely-used ones.
Correspondingly, we propose a heuristic \textit{experience elimination} mechanism based on the information density and utilization frequency of experiences, which prioritizes frequently employed experiences in task execution while discarding identified low-quality ones, streamlining the evolution of experiences toward greater efficiency.

Concretely, we measure the information gain of each shortcut by selectively identifying non-adjacent nodes whose solution optimization process exhibits an information gain exceeding a predefined threshold \(\epsilon\):
\begin{equation}
\begin{aligned}
& \bar{\mathcal{E}} = \{ (s_i, \overset{\dashrightarrow}{s_is_j}, s_j) | (s_i, \overset{\dashrightarrow}{s_is_j}, s_j) \in \mathcal{E} \\
&\qquad \wedge \omega(s_j) - \omega(s_i) \geq \epsilon \} \\
& \omega(s_j) \!=\!sim(s_j, task) \cdot sim(s_j, s_{|N|}) \cdot [\![s_j]\!] \\
\end{aligned}
\end{equation}
where \( sim(\cdot,\cdot) \) calculates the similarity between a solution with another node or a task requirement, utilizing cosine similarity and external embedders, \( [\![\cdot]\!] \) indicates a binary signal indicating whether compilation is successful via an external compiler.

Additionally, we observe a long-tail distribution in the dynamic utilization of the experience pool,  implying that the tail is in fact rarely used. 
Utilizing the frequency distribution of the experience pool utilized for each batch, we selectively eliminate certain experiences to obtain a subset with relatively high retrieval probability:
\begin{equation}
\hat{\mathcal{E}} = \{ e | e \in \mathcal{E} \wedge  \sum_{j=1}^{rank(\mathcal{E})} \frac{f(e)}{\sum_{e \in \mathcal{E}} f(e)} \le \theta \}
\end{equation}
where \( f(e) \) represents the retrieval frequency of \( e \), \( rank(\cdot) \) denotes the index of retrieval frequencies sorted in descending order, and $\theta$ represents a fractile threshold.

By combining these criteria, which consider both static information gain and dynamic usage frequency, we take the union of both to ensure the preservation of high-quality experiences:
\begin{equation}
\begin{aligned}
&\mathcal{E}_{1} = \emptyset, \quad \mathcal{E}_{2} = \mu(\bar{\mathcal{E}}_1,\mathcal{T}_2) \\
&\mathcal{E}_{i} = \mu( \bar{\mathcal{E}}_{i-1} \cup \hat{\mathcal{E}}_{i-2},\mathcal{T}_i)
\end{aligned}
\end{equation}
Note that the information-gain-based elimination can be directly processed on one generation of task batch, but the retrieval-based one requires the participation of at least two generations of task batches.
This heuristic mechanism prioritizes high-quality and frequently-utilized experiences, thereby mitigating potential inefficiencies arising from the potential expansion of the experience space.

\vspace{-6pt}
\section{Evaluation}

\vspace{-10pt}
\noindent \paragraph{Baselines} Our selection encompasses several representative and powerful LLM-driven agent frameworks specially-designed for software development.
GPT-Engineer~\cite{GPTEngineer} stands as a pioneering endeavor, harnessing the capabilities of an LLM-powered agent to engage in intricate multi-step reasoning. This innovation transcends the conventional bounds of code-level generation without the perception of past experiences, ascending to the realm of repository-level generation and realizing automated software engineering.
MetaGPT~\cite{hong2023metagpt} elevates from a single-agent pattern to a multi-agent design. Within this conceptualization, individual agents assume diverse roles akin to employees within a virtual software company. The software generation task is divided into different sub-tasks, each adeptly undertaken by an agent bespoke to a predefined role.
ChatDev~\cite{qian2023communicative} is a powerful multi-agent collaborative software development framework achieved via agent communication. ChatDev navigates the intricacies of sub-task resolution through communication established between two autonomous agents. In the procedural sequence, an instructive agent describes the details of a sub-task, such as design or code review, subsequently scrutinizing and soliciting refinements from an assistant agent, which iteratively refines the outcome in alignment with received instructions.
ECL~\cite{qian2023experiential} introduces the historical experiences into the software-developing agents. This entails the incorporation of experiential shortcuts derived from the agents' historical trajectory, which are extracted from prior software generation endeavors. These shortcuts, encapsulating distilled insights, are utilized for effectively boosting future task execution.

\vspace{-6pt}
\noindent \paragraph{Datasets} Following ECL~\cite{qian2023experiential}, we evaluate the quality of generated software on the SRDD dataset~\cite{qian2023communicative}, which contains 1,200 software requirement descriptions from 40 common categories. We perform hierarchical sampling on the dataset by software category and divide the dataset into 6 task batches. It ensures that the software description of each batch is independent and identically distributed, and has the same software category distribution. 

\vspace{-6pt}
\noindent \paragraph{Metrics} Evaluating software presents a significant challenge, particularly when aiming for a comprehensive evaluation. Traditional metrics such as function-level code assessment (e.g., \texttt{pass@k}) do not seamlessly extend to evaluating entire software systems comprehensively. This challenge primarily arises from the difficulty in creating manual or automated test cases for much of the software, especially in scenarios frequent communications, involving complex interfaces, or non-deterministic feedback. To solve this challenge, following~\citet{qian2023experiential}, we adopt three quantifiable and objective dimensions to evaluate specific aspects of the software, along with a comprehensive metric to conduct a more holistic evaluation:
\begin{enumerate}[$\bullet$]
\setlength{\parsep}{0pt}
\setlength{\topsep}{0pt}
\setlength{\itemsep}{0pt}
\item \textit{Completeness} ($\alpha$) assesses the extent of code completion in software development, calculated as the proportion of software devoid of "\texttt{TODO}" code snippets. A greater score denotes a higher degree of software completion. 
\item \textit{Executability} ($\beta$) evaluates the software's capability to operate correctly in a compilation environment, measured as the proportion of software that compiles successfully and executes directly. A greater score denotes that the software has a higher probability to execute successfully.
\item \textit{Consistency} ($\gamma$) assesses the consistency between the developed software and the initial natural language requirements. measured as the cosine distance between the embeddings of the textual requirements and the source code. A greater score denotes a closer alignment with the original requirements.
\item \textit{Quality} ($\alpha\!\times\!\beta\!\times\!\gamma$) serves as a comprehensive metric combining completeness, executability, and consistency to evaluate the quality of software comprehensively. A greater quality score indicates better overall software quality, reducing the necessity for additional manual intervention.
\end{enumerate}

\vspace{-10pt}
\noindent \paragraph{Implementation Details} We use \texttt{ChatGPT-3.5} as the foundational models. In experience acquisition, we utilize \texttt{text-embedding-ada-002} for text and code embeddings. The thresholds for experience elimination are set at $\epsilon=$0.95 and $\theta=$0.95. 
In the utilization of experience, the key-value knowledge is used through vector-based retrieval.
To ensure comparability, all other hyperparameters and environmental configurations remain identical across all baselines and our approach.

\begin{table*}[thb]
\centering
\begin{tabular}{c|cccc|c}
\toprule[1.5pt]
\textbf{} & \textbf{Completeness} & \textbf{Executability} & \textbf{Consistency} & \textbf{Quality} & \textbf{Duration} \\ 
\midrule[1.5pt]
GPTEngineer &0.4824 &0.3583 &0.7887&0.1363&15.6000  \\
MetaGPT     &0.4472 & 0.4208&0.7649&0.1439& 154.0000  \\
ChatDev     &0.7337&0.8040 &0.7909&0.4665&148.2150  \\
ECL &0.8442&0.8643 &0.7915&0.5775&122.7750 \\ \hline
IER-Successive         &\textbf{0.8744}&\underline{0.9146}&\underline{0.7968}&\textbf{0.6372}& 179.4437 \\
IER-Cumulative         &\underline{0.8492}&\textbf{0.9347}&\textbf{0.7983}&\underline{0.6337}& 181.5961 \\
\bottomrule[1.5pt]
\end{tabular}
\caption{The average performance across all methods. The \textbf{highest} scores are indicated in bold, while the \underline{second-highest} scores are underlined.}
\label{table:framework_evaluation}
\end{table*}

\vspace{-6pt}
\subsection{Quality Analysis}
We begin by assessing the software generation quality of our IER and other baselines. As shown in Table \ref{table:framework_evaluation}, there is a significant improvement over inexperienced methods such as GPTEngineer, MetaGPT, and ChatDev, evident across all quality-related metrics. Additionally, compared to ECL, which also incorporates a shortcut-oriented experience mechanism, both successive and cumulative patterns demonstrate up to an 11\% relative improvement, with the former slightly surpassing the latter. Furthermore, the average duration of IER software manufacturing does not substantially increase compared to the baselines, indicating that it does not cause excessive time delays, partly attributed to the efficiency of the vector-based retrieval design.
The advancement facilitated by IER can be attributed to its key strengths: 
1) This method enables the bypassing of certain low-level code errors and implementation issues through shortcut thinking. This allows agents to concentrate more on reviewing and optimizing intrinsic code-related problems rather than superficial implementations, ultimately enhancing software quality.
2) Besides, shortcut-oriented experiences offer more precise and detailed instructions and solutions at each optimization through their communication. This guidance directs software-developing agents to produce code with greater completeness, executability, and consistency, thereby reducing excessive delays.
3) Crucially, experience refinement through iterative propagation and elimination ensures the retention of high-quality experiences while eliminating low-quality ones, which makes the experiences more adaptive for continuous task-solving scenarios.

\begin{figure*}[t]
    \centering
    \includegraphics[width=0.99\textwidth]{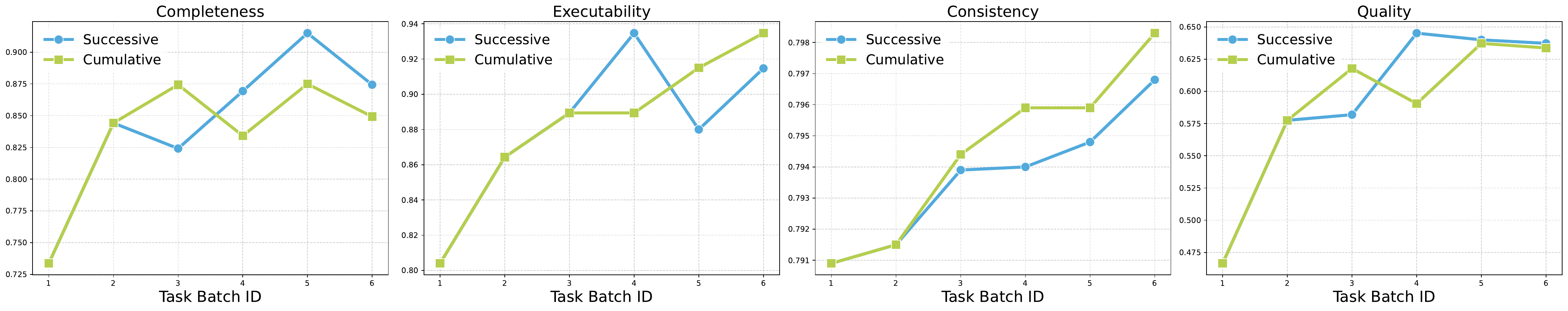}
    \caption{The average performance for each task batch across various dimensions.}
    \label{fig:generation_evaluation_quality}
    \vspace{-6pt}
\end{figure*}

\begin{figure*}[t]
    \centering
    \includegraphics[width=0.99\textwidth]{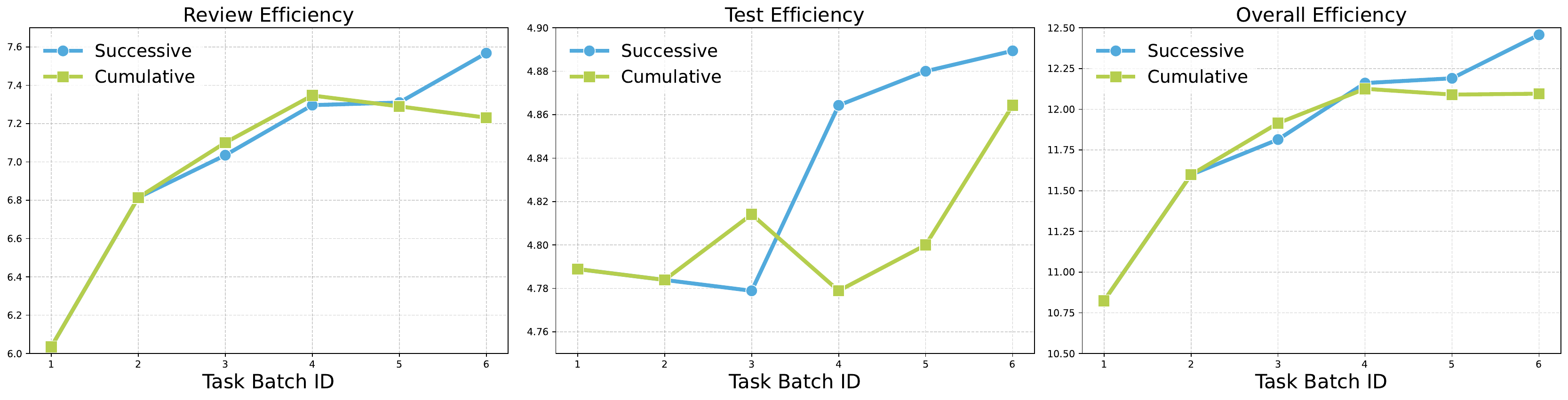}
    \caption{The phase efficiency per task batch across various dimensions. \textit{Review Efficiency} is calculated by averaging the rounds of code review, derived from the difference between the actual and maximum review rounds conducted by agents. \textit{Test Efficiency} measures efficiency during testing, while \textit{Overall Efficiency} accounts for all interactive rounds across phases, reflecting agents' whole-process software optimization. Higher results indicate faster adherence to software standards, reducing the necessity for additional manual involvement and thereby enhancing software generation efficiency.}
    \label{fig:generation_evaluation_efficiency}
    \vspace{-10pt}
\end{figure*}

\vspace{-6pt}
\subsection{Propagation Analysis}
In addition to assessing effectiveness of software quality, we show that agents equipped with high-quality experiences play a crucial role in enhancing the efficiency of software production. 
Please note that our method explicitly divides the software development process into coding, reviewing, and testing phases.\footnote{The coding phase involves a single round of agents' cooperative communication, while the reviewing and testing phases each entail multiple rounds.}
Meanwhile, all the tasks of the dataset are split into 6 disjoint batches.
Here, we examine cross-batch efficiency in the software generation process over different phases.

\paragraph{Pattern Comparison}
Figure \ref{fig:generation_evaluation_quality} and Figure \ref{fig:generation_evaluation_efficiency} illustrate the quality and efficiency results under two different refinement patterns. 
Initially, both cumulative and successive patterns yield identical results in the first and second batches. This is because the first batch doesn't use experience, and the second batch solely relies on experience propagated from the first, resulting in no discernible difference between the two patterns.
Furthermore, both patterns show a noticeable upward trend over subsequent batches, which verifies the effectiveness of our proposed iterative refinement paradigm.
Interestingly, as experience is continuously propagated among different task batches, the quality and efficiency of software manufacturing consistently improve.
The cumulative pattern exhibits a more stable trend compared to the successive pattern. This stability stems from its experience pool containing experiences from all previous batches, leading to less drastic changes with each iteration. While the successive pattern may achieve a higher upper bound in quality and efficiency, its experience pool scope remains limited to experiences from the previous batch. 
On the contrary, constant refinement of the entire experience pool in the successive pattern introduces the risk of instability. Poor experience refinements in certain batches can adversely affect the entire experience pool.

\begin{figure}[t]
    \centering
    \includegraphics[width=0.48\textwidth]{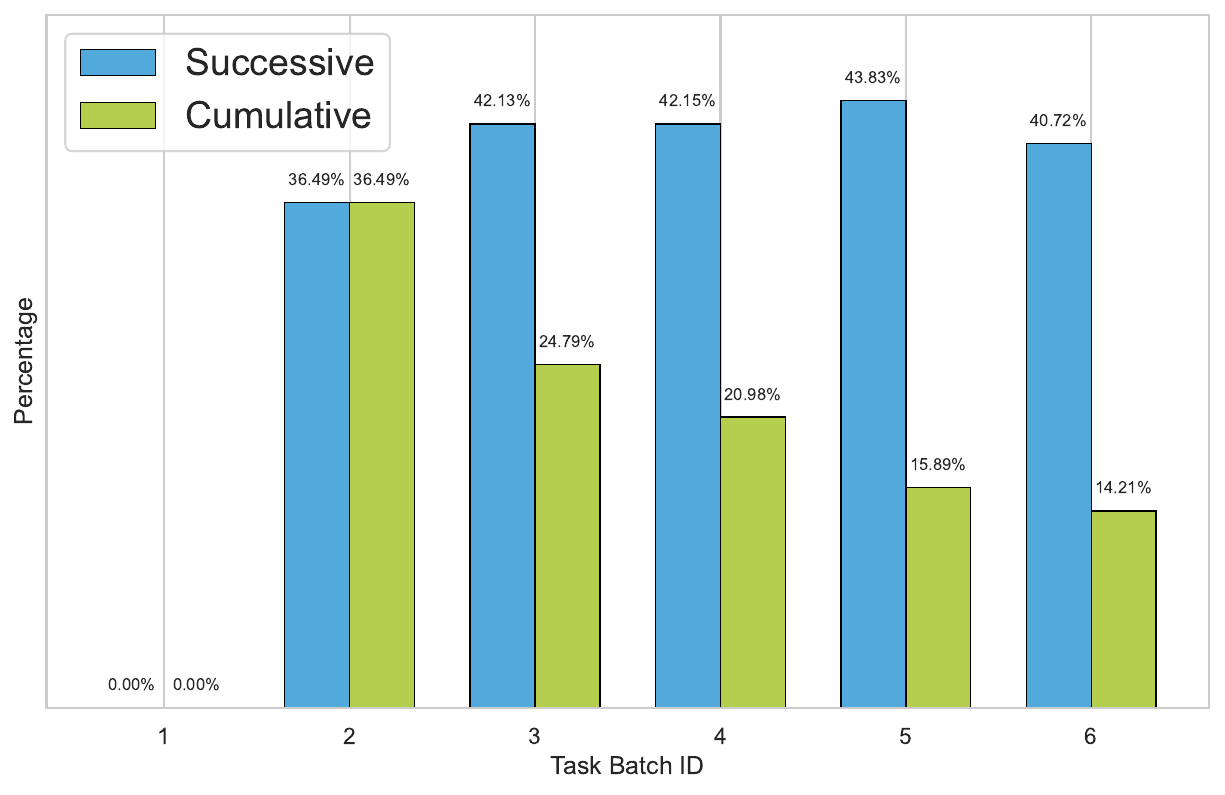}
    \caption{The retrieval hit ratio across different task batches, calculated by dividing the number of experiences retrieved by the total number of experiences.}
    \label{fig:retrieval_hit_ratio}
\end{figure}

\vspace{-10pt}
\paragraph{Utilization Analysis}
Figure \ref{fig:retrieval_hit_ratio} illustrates the fluctuation in experience retrieval hit rates across batches for the two patterns. Notably, distinct trends can be observed between the successive and cumulative patterns.
In the successive pattern, a steady increase in the hit ratio is observed, reaching its peak in the fifth batch. This trend highlights the incremental improvement in experience quality with each iteration, enabling greater utilization of experiences by descendants. However, it's important to note that experience quality may stabilize after surpassing a certain threshold, leading to stabilization rather than continuous improvement.
In contrast, the cumulative pattern exhibits a gradual decline in the hit ratio, indicating a degradation in propagated experience quality due to the exponential growth of the experience pool, which inevitably includes numerous low-quality or rarely-used ones. 
This phenomenon underscores the urgent need for experience elimination, especially for the cumulative pattern, aligning with the intuitive motivation for proposing this crucial mechanism.

\begin{figure}[t]
    \centering
    \includegraphics[width=0.48\textwidth]{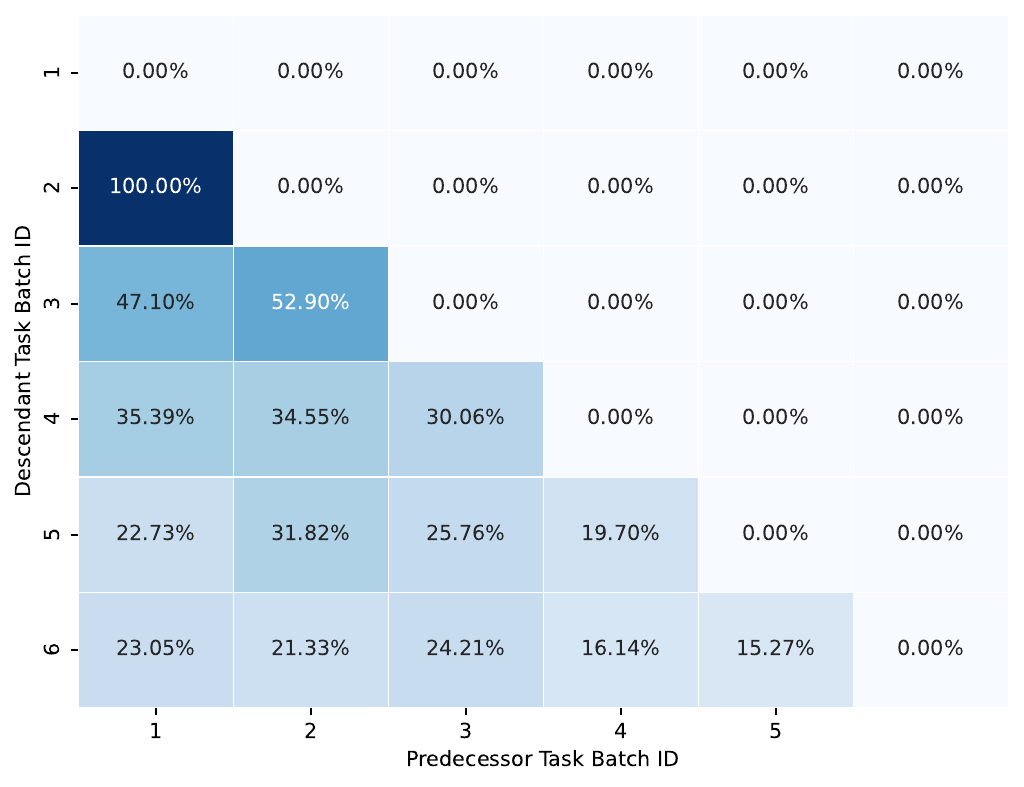}
    \caption{The distribution of the cumulative pattern across all task batches. Please note that the distribution of the successive pattern is not depicted, as it only shows a single diagonal line resembling an identity matrix.}
    \label{fig:inheritance_map}
\end{figure}

\noindent \paragraph{Utilization Distribution}
Having explored the retrieval-based utilization of the experience pool by descendants, we now analyze the overall distribution of experiences utilized by one batch from all its predecessors, resulting in the utilization distribution depicted in Figure \ref{fig:inheritance_map}.
Our findings regarding the utilization distribution in the cumulative pattern are summarized as follows:
1) Experiences obtained from a predecessor are utilized by all descendants, not only the nearby one.
2) Vertically, there is a decline in each column from top to bottom, suggesting a reduction in the utilization frequency of experiences produced by more distant descendants.
3) Horizontally, experiences acquired by a descendant are not mainly derived from its nearest predecessor but are distributed approximately uniformly, highlighting that experiences propagated from different predecessors remain relevant.

\begin{figure}[t]
    \centering
    \includegraphics[width=0.48\textwidth]{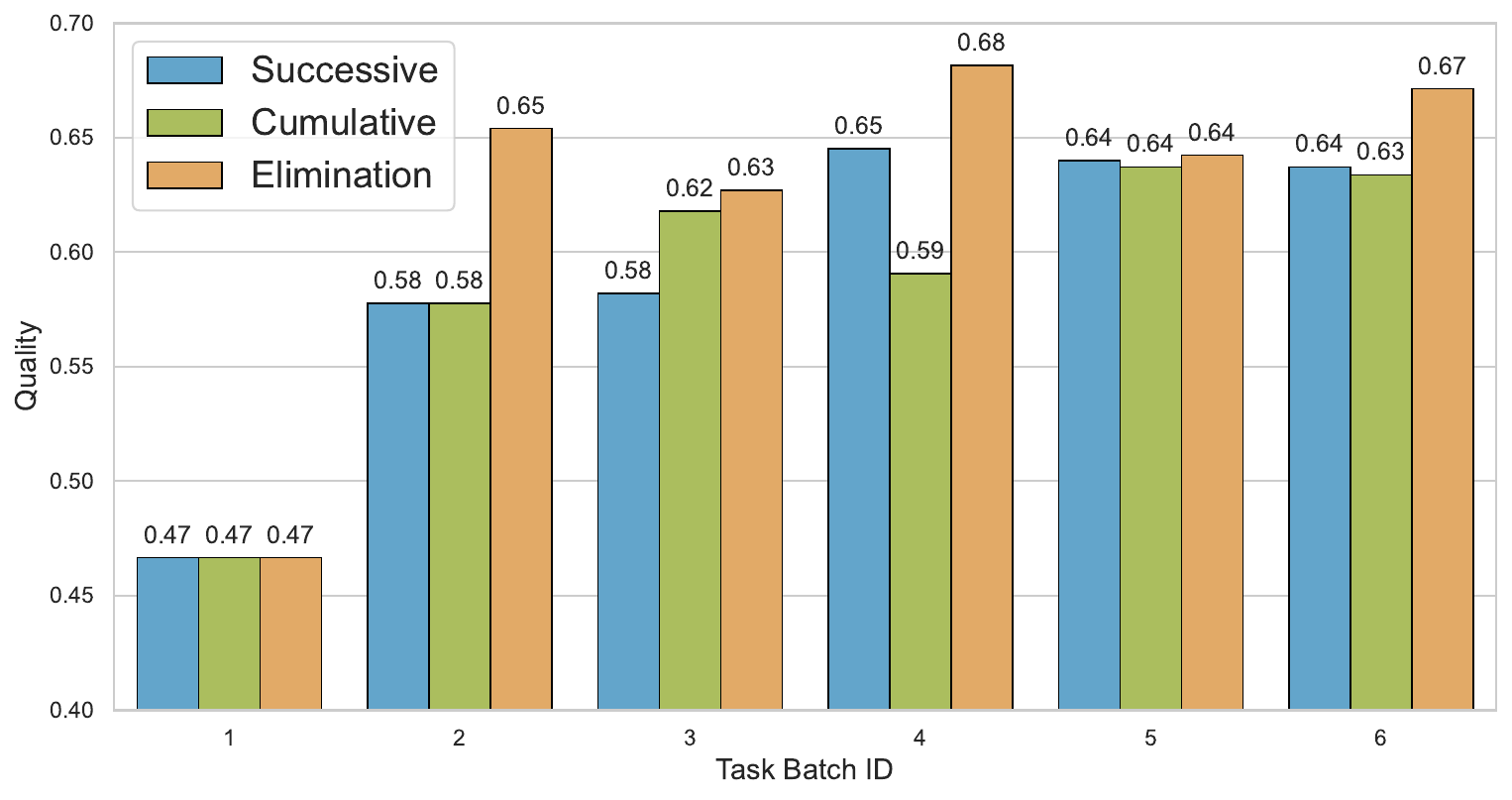}
    \caption{Comparison of software quality between the fundamental patterns and the variant enhanced with experience elimination.}
    \label{fig:filter_quality}
    \vspace{-10pt}
\end{figure}

\subsection{Elimination Analysis}
We have shown that the cumulative pattern provides a more stable utilization of experiences, potentially recalling a broader range of historical experiences from all predecessors. As the pool size continuously expands, high-quality experiences in the pattern inevitably become diluted across the entire experience pool, resulting in a long-tail distribution of experience utilization.
As shown in Figure \ref{fig:filter_quality}, the elimination mechanism guarantees the concentration of high-quality experiences in the pool, resulting in comparable or even superior quality metrics across all batches. Empirically, the mechanism in our setting utilizes only 11.54\% of the experience pool size, compared to the non-eliminated one, resulting in a total of 930 experiences after elimination, down from 8053 initially. This naturally strikes a trade-off between the volume and the utilization of experiences, making it highly recommended for application in real-world systems.

\vspace{-6pt}
\section{Conclusion}
We've introduced an iterative experience refinement framework, enabling LLM agents to refine experiences iteratively during continual task execution. We proposed both the successive and cumulative patterns for experience refinement, alongside a heuristic experience elimination mechanism to effectively manage the experience space while enhancing performance.
Our experiments show that while the successive pattern may yield higher performance, the cumulative pattern provides more stable performance. Additionally, experience elimination allows achieving superior performance using only 11.54\% of a high-quality subset.
We anticipate that our insights will catalyze a paradigm shift in shaping the design of LLM agents, driving them towards greater autonomy and fostering evolutionary growth in collective intelligence.

\clearpage
\bibliography{references}

\end{document}